\documentclass[sn-mathphys-num]{sn-jnl}

\usepackage{graphicx}%
\usepackage{multirow}%
\usepackage{amsmath,amssymb,amsfonts}%
\usepackage{amsthm}%
\usepackage{mathrsfs}%
\usepackage[title]{appendix}%
\usepackage{xcolor}%
\usepackage{textcomp}%
\usepackage{manyfoot}%
\usepackage{booktabs}%
\usepackage{algorithm}%
\usepackage{algorithmicx}%
\usepackage{algpseudocode}%
\usepackage{listings}%

\usepackage{mathtools}
\usepackage{array}

\theoremstyle{thmstyleone}%
\newtheorem{theorem}{Theorem}
\theoremstyle{thmstyletwo}%

\theoremstyle{thmstylethree}%
\newtheorem{definition}{Definition}%

\newtheorem{lemma}{Lemma}

\newcommand{\norm}[1]{ \lVert {#1} \rVert}
\newcommand{\card}[1]{ \left| {#1} \right| }
\newcommand{\prts}[1]{ \left( {#1} \right) }
\newcommand{\brkt}[1]{ \left\{ {#1} \right\} }
\newcommand{\sqbrkt}[1]{ \left[ {#1} \right] }
\newcommand{\argmax}[1]{\mathop{\mathrm{argmax}}\nolimits_{#1}}

\usepackage[draft,inline,nomargin]{fixme} \fxsetup{theme=color}
\FXRegisterAuthor{cg}{cgg}{\color{blue}CG}

\begin{document}

\title{\begin{center}
The Art of Misclassification: \\ Too Many Classes, Not Enough Points
\end{center}}


\author*[1]{\fnm{Mario} \sur{Franco}}\email{mfrancomndez@binghamton.edu}
\author[1,2]{\fnm{Gerardo} \sur{Febres}}
\author[1,3]{\fnm{Nelson} \sur{Fernández}}
\author[1]{\fnm{Carlos} \sur{Gershenson}}


\affil*[1]{\orgname{School of Systems Science and Industrial Enginnering. Binghamton University}, \orgaddress{\city{Binghamton}, \country{USA}}}
\affil[2]{\orgname{Universidad Simón Bolívar}, \orgaddress{\city{Caracas}, \country{Venezuela}}}
\affil[3]{\orgname{Grupo de Investigación en Ecología y Biogeografía. Universidad de Pamplona}, \orgaddress{\city{Pamplona}, \country{Colombia}}}


\abstract{
    Classification is a ubiquitous and fundamental problem in artificial intelligence and machine learning, with extensive efforts dedicated to developing more powerful classifiers and larger datasets. 
    However, the classification task is ultimately constrained by the intrinsic properties of datasets, independently of computational power or model complexity. 
    In this work, we introduce a formal entropy-based measure of classificability, which quantifies the inherent difficulty of a classification problem by assessing the uncertainty in class assignments given feature representations. 
    This measure captures the degree of class overlap and aligns with human intuition, serving as an upper bound on classification performance for classification problems. 
    Our results establish a theoretical limit beyond which no classifier can improve the classification accuracy, regardless of the architecture or amount of data, in a given problem. 
    Our approach provides a principled framework for understanding when classification is inherently fallible and fundamentally ambiguous.
}





\maketitle

\section{Introduction}\label{sec1}

Since the origins of life, the struggle to classify has reached every nook and cranny of this planet. 
Bacteria need to differentiate food from predators, neurons' excitatory signals from inhibitory signals, dogs' tasty chicken from boring kibble, and humans' images of bicycles from pictures of traffic lights to be officially recognized as humans. 
Classification is a ubiquitous task for any system capable of action that has a preference over its environment and itself. 
It holds such importance for humans that considerable efforts have been dedicated to developing strong and automated classification methods. 

\begin{figure}[!th]
    \centering
    \includegraphics[width=0.3\textwidth]{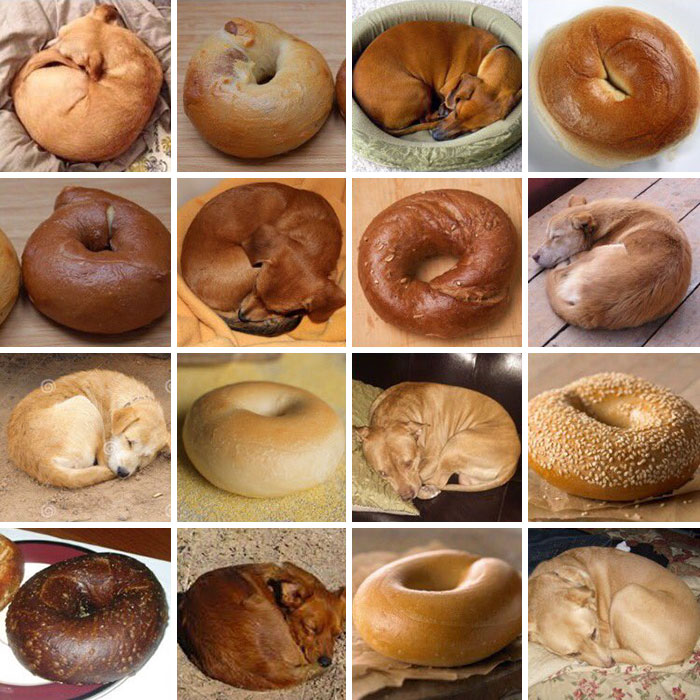}
	\includegraphics[width=0.3\textwidth]{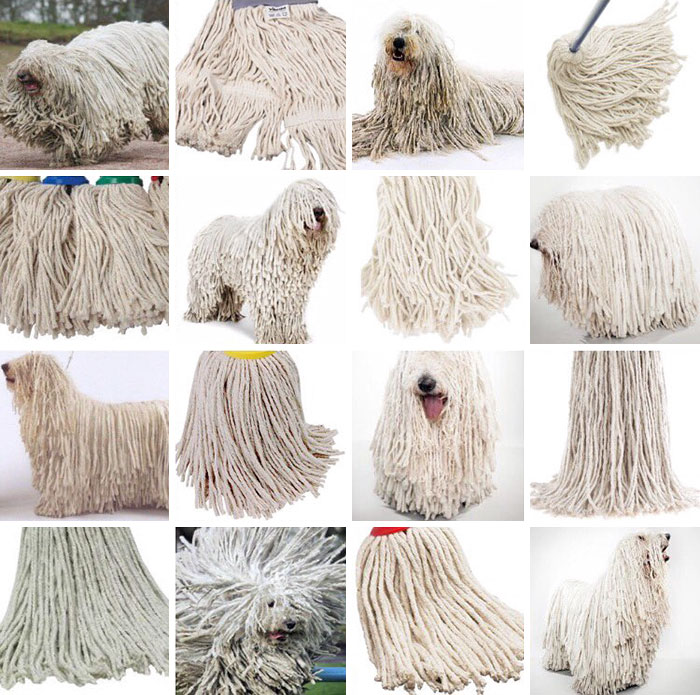}
    \includegraphics[width=0.3\textwidth]{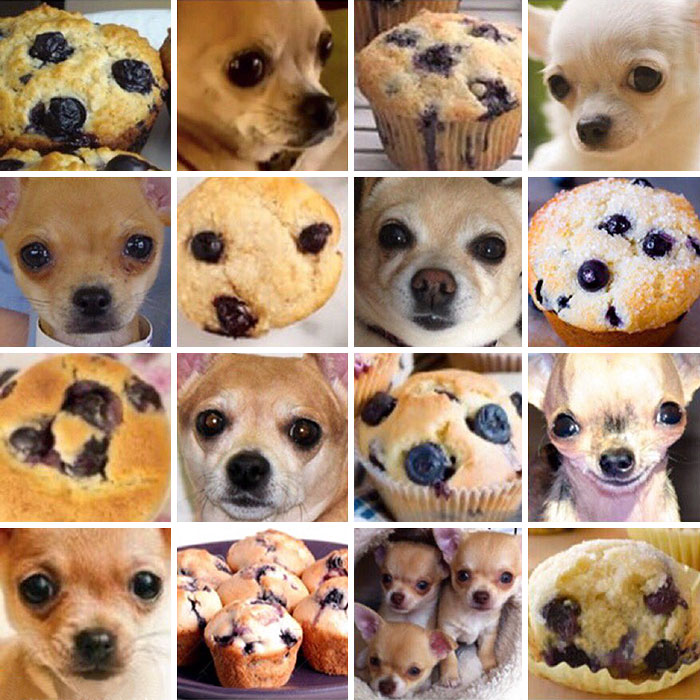}
    \caption{An example of a challenging classification problem. Even though we consider dogs to be in a completely different class than breads and mops, in some cases, their resemblance can be uncanny.}
    \label{fig:dogs}
\end{figure}

In an economic context, accurately assessing the situation can mean the difference between success and bankruptcy; in healthcare, it can mean the difference between life and death.  
This latter scenario has become particularly relevant with the rise of computers and machine learning, especially deep learning \cite{healthcare}. 
However, an often overlooked fact about classification is that classes are not natural features but rather artificial attributes that we (somewhat arbitrarily) assign to certain phenomena.  
Such phenomena, which may be clearly defined on their own, may overlap with others that we consider simultaneously, given our world observations. 
This extends far beyond distinguishing between puppy or bagel, sheepdog or mop, or the famous challenge of chihuahuas and blueberry muffins (see Figure~\ref{fig:dogs}) \cite{chihuahua_blueberry}.

Constructing a suitable classifier can be challenging, depending on the degree of overlap.  
As a result, many classification problems have an intrinsic limitation that cannot be overcome by allocating more computational resources and data points.  
In other words, there is an inherent aspect of the observations that induces a hard bound on the classification problem.  
Although the term 'classificability' is not widely standardized in the literature, it is closely related to the concept of separability, which describes how distinctly different classes can be identified within a given representation space. 
However, separability alone does not fully capture the inherent challenges of classification problems. 

While much of the research in machine learning focuses on improving classifiers, a growing body of work investigates the fundamental constraints on classification or separation performance. 
For example, in~\cite{separability_measure_guan}, the authors use the Kolmogorov-Smirnov similarity of the distribution of distances of the classes to assess the separability of several datasets.
Fisher's Discriminant Ratio~\cite{common_discriminant} is another method to evaluate the separability of a dataset that relies on the statistics of the samples.
In the same work~\cite{common_discriminant}, the authors also define other metrics based on the overlap of bounding boxes and the ability of individual features to separate the samples.  
Other approaches rely on building minimum spanning trees for the samples and measure the frontiers between nodes of different classes~\cite{mst_multivariate, common_discriminant}.
Perhaps a more natural measure is to use topological neighborhoods of the samples to construct a separability measure~\cite{topological_neighborhoods, common_discriminant}.
For an extensive review of existing methods, we refer the reader to a more comprehensive analysis such as~\cite{measures_review}.

Nevertheless, we believe some of these metrics do not always predict this limit or assume specific properties about the data that may not always hold in practice. 
In contrast, incorporating information-theoretic perspectives can provide an alternative approach that allows us to get closer to capturing the intrinsic limits of classification across complex datasets. 
In this sense, incorporating information-theoretic perspectives can provide an alternative approach that brings us closer to capturing the inherent limits of classification. 
Building on this idea, we propose an entropy-based measure that provides a principled way to estimate an upper bound on classification performance. 
This measure serves as a theoretical limit beyond which no classifier can improve the classification accuracy, regardless of the architecture or amount of data, in a given problem. 

This work is structured as follows: Section 2 illustrates the importance of estimating the limit of a classification problem. 
Section 3 introduces several definitions and the main equation used for the limit estimation. 
Section 4 translates the main equation into a workable definition and provides a simple classification example.
Section 5 applies the methodology to more sophisticated synthetic problems to further illustrate the potential of the methodology.
Section 6 provides a simple summary of the effect of the problem dimensionality of the limit estimation.
Section 7 applies this methodology to several real-world problems to showcase the utility of this concept in practice.
Finally, section 8 provides conclusions and future directions of the present work.
Detailed code for the experiments is available on GitHub\footnote{https://github.com/Nogarx/the-art-of-misclassification}.

\section{Why does it matter?}\label{sec2}

As mentioned above, several efforts have been put into developing bigger, fancier,  and more powerful classifiers. 
The question then is: why invest time in a measure of classificability instead of building an almighty classifier?
It turns out that, in practice, perfect accuracy may be an illusion, either the product of our powerful classifier or a lack of enough evidence of the underlying distribution.
For example, it does not matter how big of a model one uses to try to predict a truly stochastic coin toss; the best we can do is guess with the same bias that the coin toss has.

\begin{figure}[!t]
    \centering
    \begin{minipage}[]{0.75\textwidth}
        \includegraphics[width=1\textwidth]{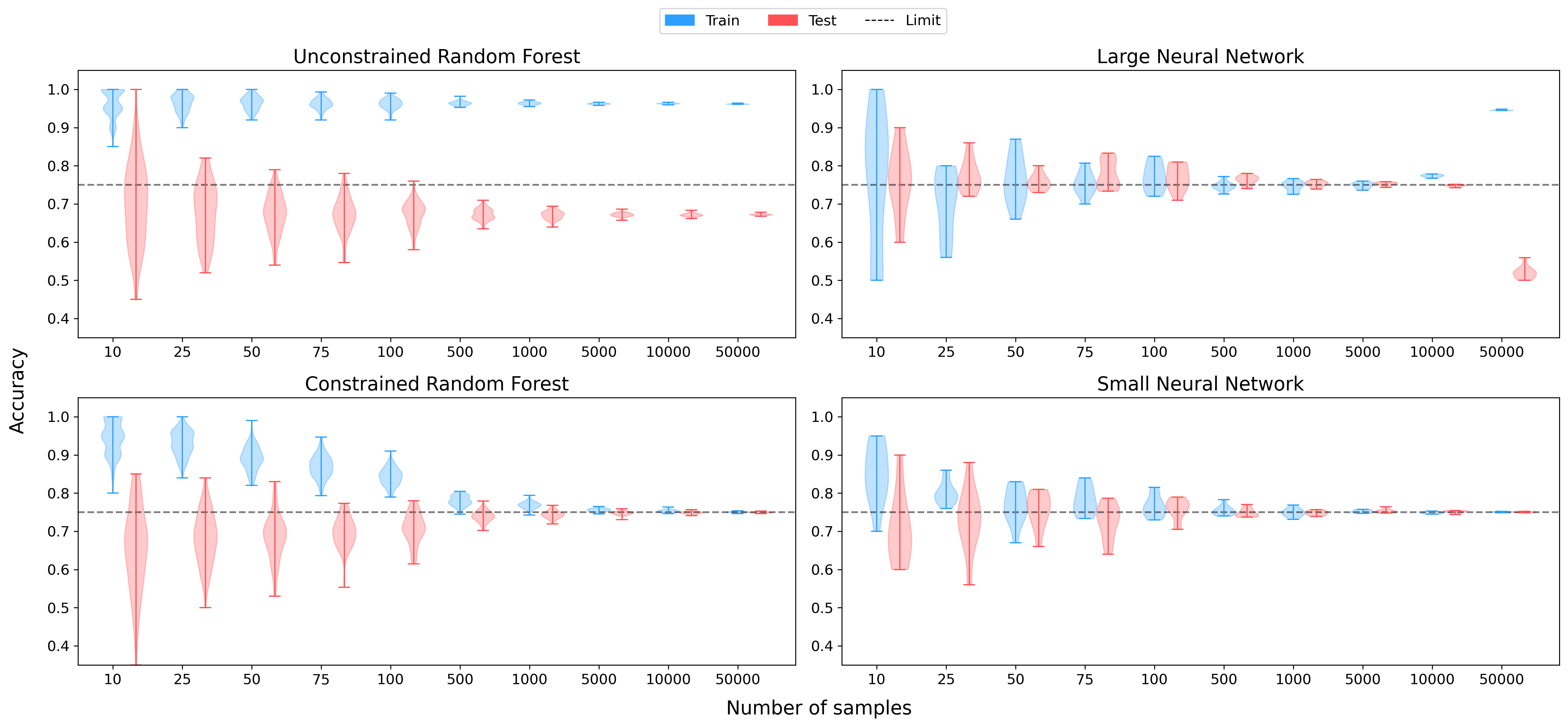}
    \end{minipage}
    \begin{minipage}[]{0.24\textwidth}
        \includegraphics[width=1\textwidth]{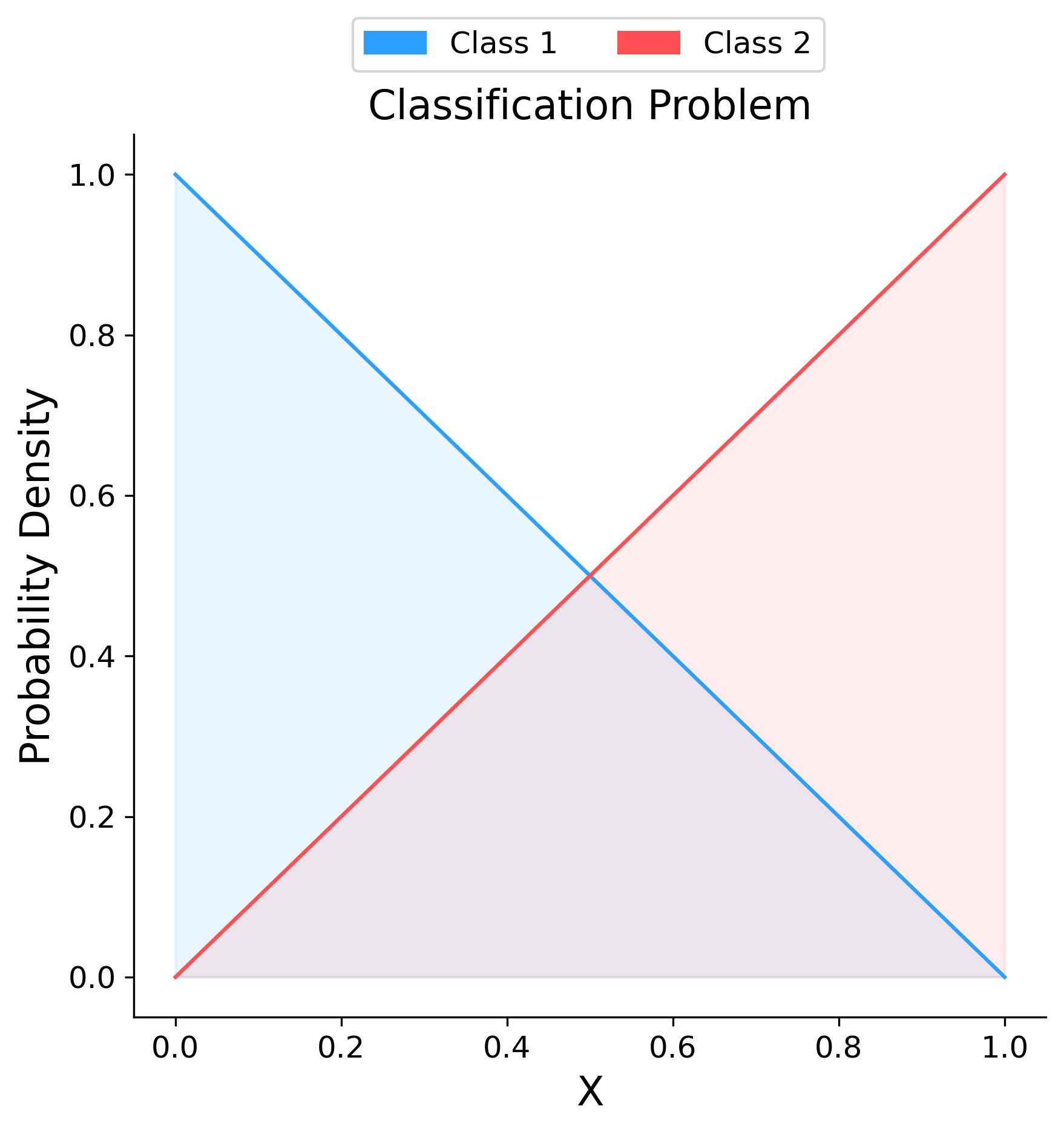}
    \end{minipage}
    \caption{
    Accuracy of an unconstrained random forest (top left), a constrained random forest (bottom left), a large neural network (top right), and a small neural network (bottom right) as a function of the number of samples. 
    This problem's theoretical classificability limit is shown as a dashed black line.
    The classificability limit is an intrinsic upper bound for a simple classification task of two classes with probability distribution $x$ and $1-x$ restricted to the domain $\sqbrkt{0,1}$.
    To the right of the figure, we showcase the distribution of each class used to construct the datasets.    
    Note that the unconstrained random forest and the large neural network classifiers perform worse than their constrained/smaller version, even though, in theory, they are more expressive models, and both observe large amounts of data.
    For neural networks, we took the accuracies at the maximum accuracy value observed at any epoch on the test partition.
    The large neural network is a 6-layer network (1024, 1024, 512, 512, 256, 128 units); the small neural network is a 3-layer (128, 64, 32 units).
    Both neural networks were trained with the Adam optimizer without regularization.
    }
    \label{fig:why_example}
\end{figure}

As a simple example, consider how the accuracy of the two most popular classifiers, random forest and neural networks, change as a function of the number of samples and the expressiveness of the model for an extremely simple dataset with two classes on a single variable (Figure~\ref{fig:why_example}). 
Intuition would tell us that the more expressive a model is, the better it will perform as we increase the number of samples.
This intuition is often referred to as the Scaling Hypothesis, or in the context of neural networks, as Neural Scaling Laws~\cite{neural_scaling}.
In practice, this is only sometimes the case; a more expressive model can fail to learn a straightforward problem, even with large amounts of data.
Perhaps more interestingly, sometimes a model can manage to properly learn the underlying distribution with a relatively small amount of data. 
Still, as soon as we ramp up the amount of data, the same model can fail to generalize, as we can see with the large neural network.
Conversely, simpler models can converge near an optimal solution with less data and even generalize well for large amounts of data.
This rather simple observation challenges the na\"ive intuition that bigger is always better. 

\section{The Entropy Limit}\label{sec3}

The previous section highlighted the distinction between a dataset and a classification problem: a dataset is a finite realization of a classification problem.
Frequently, these two terms are confused by machine learning practitioners, and it is common to encounter dataset-centric competitions when what we actually want is a problem-centric solution.
Thus, before continuing any further, it is helpful to formalize what we mean by a class, a dataset, and a classification problem. 

\medskip
\begin{definition}
We define a \textbf{\textit{class}}, $\alpha$, as a set of random variables $\brkt{X_{i}}$ with sample spaces $\Omega_{i}$.
We denote by $f_{\alpha}(\bar{x}) = f_{X_{0}, \cdots, X_{n}}(x_{0}, \cdots, x_{n})$ the probability densitiy function of the class $\alpha$ and $\Omega_{\alpha} = \Omega_{0} \times \cdots \times \Omega_{n}$ the sample space of $\alpha$.
\end{definition}
\medskip

Note that it is always possible to extend the domain of $\alpha$ with the introduction of another random variable $X_{j}$ via a simple product of the domains $\Omega_{\alpha} \times \Omega_{j}$ such that the extension $f_{\alpha}^{*}\prts{x_{0}, \cdots, x_{i}, x_{j}}=f_{\alpha}\prts{x_{0}, \cdots, x_{i}}$, i.e., the extension is independent of the variable $X_{j}$, and $f_{\alpha}^{*}$ is appropriately normalized. This extension is handy when we want to form a dataset with classes $\alpha$ and $\beta$ that do not have compatible sample spaces. 

\medskip
\begin{definition}
Let $\Gamma$ be a set of classes with associated density functions $f_{\gamma}$, we define a \textbf{\textit{classifiable dataset}}, $D$, as a finite set of tuples  $\brkt{ \: (\bar{x}, \alpha) \: \mid \: \alpha \in \Gamma, \: \bar{x} \in \Omega, \: \bar{x} \sim f_{\alpha} \:}$ with sample space $\Omega = \cup_{\alpha \in \Gamma} \: \Omega_{\alpha}$. 
\end{definition}
\medskip
\begin{definition}
Let $\Gamma$ be a set of classes with associated density functions $f_{\gamma}$ and $\mathcal{A}$ a function for the accuracy of a model, we define a \textbf{\textit{classification problem}}, $\Pi$, as finding a model $\mathcal{M}^{*}$ such that $\mathcal{M}^{*} = \argmax{\mathcal{M}} \mathbb{E} \sqbrkt{  \mathbb{E} \sqbrkt{  \mathcal{A} \prts{\mathcal{M}, D} \mid D}} = \argmax{\mathcal{M}} \mathbb{E} \sqbrkt{  \mathcal{A} \prts{\mathcal{M}, D}}$, i.e., a model that maximizes the expected accuracy over all possible datasets generated by $f_{\gamma}$.
\end{definition}
\medskip

It is also possible to give an equivalent set of definitions using the classification problem as the root instead of the concept of class, but we consider the concept of class to be more natural.
Intuitively, the previous definition formulates a classification problem in terms of finding a model that behaves well on all possible datasets.
Consequently, a good measure of classificability should find the limits of the classification problem itself, not the particular instantiations of the problem (the datasets).

Now, recall that maximizing the accuracy is an equivalent problem to minimizing the negative log-likelihood of the cross entropy.
For the sake of keeping the notation consistent with the rest of the literature, we can think of a dataset as the result of a sampling process of some probability density function $p$ and a model as some sampling process with probability distribution $q$.
Hence, $\mathcal{M}^{*}$ is the model that minimizes $\mathbb{E} \sqbrkt{ -\log \prts{q} \mid p } = H\prts{p,q}$.
Consequently, the accuracy of $\mathcal{M}^{*}$ is intrinsically related to the entropy of the classification problem $\Pi$.
Moreover, if we normalize the entropy $H$, to make it invariant to the choice of model $q$, we can obtain an upper bound for the maximum accuracy achievable for the problem $\Pi$.

\medskip
\begin{definition}
We define the classificability limit $\mathscr{C}_{\Pi}$ of a problem $\Pi$ as,
\begin{equation}
	\mathscr{C}_{\Pi} = 1 - \frac{\card{\Gamma}-1}{\card{\Gamma}} \int_{\Omega} \sum_{\alpha \in \Gamma} p \prts{ \bar{x}, \alpha } \log \prts{ \frac{ p \prts{ \bar{x}, \alpha } }{ \rho \prts{ \bar{x}, \alpha } } } \,d \bar{x}
\label{eq:classificability},
\end{equation}
where,
\begin{equation}
	p \prts{ \bar{x}, \alpha } = \frac{f_{\alpha}(\bar{x})}{\sum_{\beta \in \Gamma}f_{\beta}(\bar{x})}
\label{eq:rel_prob},
\end{equation}
and $\rho \prts{ \bar{x}, \alpha }$ is an invariant measure, such that the integral is normalized and invariant to a change of coordinates. 
\label{def:classificability}
\end{definition}
\medskip

Observe that eq. \ref{eq:rel_prob} is just the relative probability between classes and the integral of eq. \ref{eq:classificability} is the entropy of the classes.
This entropy formulation is often called the relative entropy or the Kullback–Leibler divergence \cite{kl_divergence}. 
Still, it can be shown that this is the corresponding extension of Shanon's entropy for continuous spaces~\cite{jaynes_lectures, jaynes_prior}.

This definition naturally fulfills two of our intuitions: 1) when the entropy is zero, the classes are perfectly separable, and thus the corresponding limit is equal to one; 2) when the entropy is one, the classes are completely scrambled, and any strategy will not be better than just random guessing, which will be correct with a probability $\frac{1}{\card{\Gamma}}$, assuming that every class is equally represented.


It is important to remark that the leading term of the integral is misleading.
It is not a constant but a function with an explicit spatial dependence.
We picked a particular value of such function for the sake of intuition, but this value only holds for equally represented classes of constant distributions.
However, this factor is not of significant importance in practice, as it can be absorbed by the $\rho$ term, as discussed next.



\section{Entropy Estimation}\label{sec4}

In practice, estimating any phenomenon's actual probability density function is non-trivial.
To worsen the situation, our measurements of any phenomenon tend to include instrument and observer errors.
And to go from the frying pan into the fire, to the best of our knowledge, there is no algorithm to compute the limiting density function $\rho$, even if the PDFs of every class are known.
Therefore, all hope is lost, or that would be the case until we recall that entropy is additive and the idea behind the Riemann integral.

Fortunately, it is relatively simple to compute the density function of a constant distribution.
First, consider a constant PDF with a set of classes $\Gamma$ and a domain $\Omega = \brkt{*}$.  
Then, starting from equation \ref{eq:classificability} we have that,
\begin{equation}
	\mathscr{C}_{\Pi} = 1 - K \sum_{\alpha \in \Gamma} p \prts{ \alpha } \log \prts{ \frac{ p \prts{ \alpha } }{ \rho \prts{ \alpha } } },
\end{equation}

After absorbing the $K$ factor into $\rho$ and realizing that the strategy for an optimal model in this setup is to always bet for the most common class (or, if stochasticity is preferred, betting with the same odds of observing each class), it follows that, 
\begin{equation}
	 1 - \max_{\beta} \brkt{p \prts{ \beta }}  = \sum_{\alpha \in \Gamma} p \prts{ \alpha } \log \prts{ \frac{ p \prts{ \alpha } }{ \rho \prts{ \alpha } } },
\end{equation}
From here, it is just algebraic gymnastics to obtain that, 
\begin{equation}
	 \rho \prts{ \alpha } = p \prts{ \alpha } \cdot e^{1 - \max_{\beta} \brkt{p \prts{ \beta }}} = p \prts{ \alpha } \cdot  \min_{\beta} \brkt{ e^{1 - p \prts{ \beta }}}
\end{equation}

In general, a similar result holds for constant distributions with more interesting domains; the only difference is that $rho$ now has an explicit dependence on the space.
\begin{equation}
	\rho \prts{ \bar{x}, \alpha } = p \prts{ \bar{x}, \alpha } \cdot \min_{\beta} \brkt{ e^{1 - p \prts{ \bar{x}, \beta } } }
\label{eq:constant_density}
\end{equation}

Now, we need to estimate the entropy around every point in the dataset.
To compute such an estimation, we need to know the probability of each class around each given point.
Again, this is also unfeasible in practice, but we can compute a rough estimate by using nearby points. 

Therefore, we estimate the classificability of a classification problem $\Pi$ given a dataset $D$ using the expectation of the individual entropies as follows,

\begin{equation}
	\widehat{\mathscr{C}_{D}} = 1 - \mathbb{E} \sqbrkt{ \brkt{ \mathcal{H} \prts{ \bar{x}, \theta } \mid \prts{ \bar{x}, \alpha } \in D } }  ,
\end{equation}

where,

\begin{equation}
	\mathcal{H} \prts{ \bar{x}, \theta } = 
	- \sum_{ \alpha } \hat{p} \prts{ \bar{x}, \alpha } \log \prts{ \frac{ \hat{p} \prts{ \bar{x}, \alpha } }{ \hat{\rho} \prts{ \bar{x}, \alpha } } } ,
\end{equation}

\begin{equation}
	\hat{\rho} \prts{ \bar{x}, \alpha } = \hat{p} \prts{ \bar{x}, \alpha } \cdot \min_{\beta} \brkt{ e^{1 - \hat{p} \prts{ \bar{x}, \beta } } } ,
\end{equation}

\begin{equation}
	 \hat{p} \prts{ \bar{x}, \alpha } =  \sum_{ \prts{ \bar{y}, \gamma } \in \mathcal{B} \prts{ \bar{x}, \theta } }  \frac{ \delta \prts{ \alpha - \gamma } }{ \card{\mathcal{B} \prts{ \bar{x}, \theta } } } ,
\end{equation}

\begin{equation}
	 \mathcal{B} \prts{ \bar{x}, \theta } = \brkt{ \prts{ \bar{y}, \gamma } \mid \norm{ \bar{y} - \bar{x} } < \theta , \prts{ \bar{y}, \gamma } \in D \setminus \brkt{\bar{x}}  } .
\end{equation}
\smallskip

To clarify the previous notation, note that $\mathcal{B} \prts{ \bar{x}, \theta } \subseteq D$ is simply the subset of samples that is close to the point $\bar{x}$ and the hat notation indicates that the function is a local estimation of the real functions. 
Moreover, the form of the density function for a constant distribution allows us to further simplify the entropy as a fraction of probabilities, which can be helpful for numerical estimations.

\medskip
\begin{lemma}
Let $\theta \in \mathbb{R}$ be some threshold and $D$ be some dataset composed of $k$ classes, such that $f_{\alpha}$ is the PDF associated with the class $\alpha$ and $\forall \alpha$, $p_{\alpha}$ is continuous in the domain of $D$. 
Then, the estimator $\hat{p}$ converges to $p$ as $\theta \to 0$ and $\card{D} \to \infty$. In other words,
\begin{equation}
	 p \prts{x, \alpha} = \lim_{ \begin{smallmatrix} \theta \to 0 \\ \card{D} \to \infty \end{smallmatrix} } \hat{p} \prts{x, \alpha} .
\end{equation}

When we say that $\card{D} \to \infty$, we assume all the samples are independent and identically distributed (with their respective class biases).
\begin{proof}
The proof of this lemma follows immediately from the Law of large numbers.
\end{proof}
\label{lemma:p_converges}
\end{lemma}
\medskip
\begin{lemma}
Let $\theta \in \mathbb{R}$ be some threshold and $D$ be some dataset composed of $k$ classes, such that $f_{\alpha}$ is the PDF associated with the class $\alpha$ and $\forall \alpha$, $p_{\alpha}$ is continuous in the domain of $D$. 
Then, the estimator $\hat{\rho}$ converges to $\rho$ as $\theta \to 0$ and $\card{D} \to \infty$. In other words,
\begin{equation}
	 \rho \prts{x, \alpha} = \lim_{ \begin{smallmatrix} \theta \to 0 \\ \card{D} \to \infty \end{smallmatrix} } \hat{\rho} \prts{x, \alpha} .
\end{equation}
\begin{proof}
Without loss of generality, let $\bar{x} \in \Omega$ and note that, $\forall \alpha$, $f_{\alpha}$ is a continuous function, consequently $p \prts{\bar{x}, \alpha}$ is a contiunous function.
Moreover, by definition of continuity $\exists \epsilon > 0$ such that $\forall \alpha$, $\exists \delta_{\alpha} > 0$ so that $\norm{p \prts{\bar{x}, \alpha} - p \prts{\bar{x}+\delta_{\alpha}, \alpha}} < \epsilon$.
Take $\delta = \min \brkt{\delta_{\alpha}}$.
Let $Q$ be a countable partition of $\Omega$ such that $q = \mathcal{B} \prts{ \bar{x}, \delta }$ and $q \in Q$.
Since entropy is additive, we have that,
\begin{equation}
    \mathcal{H} \prts{ D } = \sum_{r \in Q} \mathcal{H} \prts{ r } = \mathcal{H} \prts{ q } + \sum_{r \in Q \setminus \brkt{q}} \mathcal{H} \prts{ r } 
\end{equation}
Observe that $\forall y \in q$, $\norm{p \prts{\bar{x}, \alpha} - p \prts{y, \alpha}} < \epsilon$, in other words $p \prts{q, \alpha} \sim c_{\alpha}$, subject to $\sum_{\alpha} c_{\alpha} = 1$. 
Hence $\mathcal{H} \prts{ q }$ is an entropy of constant PDFs, thus $\hat{\rho} \prts{ \bar{x}, \alpha } = \hat{p} \prts{ \bar{x}, \alpha } \cdot \min_{\beta} \brkt{ e^{1 - \hat{p} \prts{ \bar{x}, \beta } } }$.
By lemma \ref{lemma:p_converges} $\hat{p} \to p$ as $\card{D} \to \infty$ so,
\begin{equation}
	\lim_{ \begin{smallmatrix} \theta \to 0 \\ \card{D} \to \infty \end{smallmatrix} } \hat{\rho} \prts{\bar{x}, \alpha} = p \prts{ \bar{x}, \alpha } \cdot \min_{\beta} \brkt{ e^{1 - p \prts{ \bar{x}, \beta } } }
\end{equation}

However, the previous equation precisely defines $\rho$ for a constant PDF. 

Finally, this holds $\forall \bar{x} \in \Omega$; therefore, $\rho$ can be approximated locally as it if were a constant distribution.  
\end{proof}
\label{lemma:rho_converges}
\end{lemma}
\medskip
\begin{theorem}
Let $D$ be some dataset and $\theta \in \mathbb{R}$ some threshold. Then, the estimator $\widehat{\mathscr{C}_{D}}$ converges to the true ${\mathscr{C}_{D}}$ as $\theta \to 0$ and $\card{D} \to \infty$. In other words,
\begin{equation}
	 \mathscr{C}_{D} = \lim_{ \begin{smallmatrix} \theta \to 0 \\ \card{D} \to \infty \end{smallmatrix} } 1 - \mathbb{E} \sqbrkt{ \brkt{ \mathcal{H} \prts{ \bar{x}, \theta } \mid \prts{ \bar{x}, \alpha } \in D } }.
\end{equation}
\begin{proof}
The proof follows immediately from the lemmas \ref{lemma:p_converges} and \ref{lemma:rho_converges}.
\end{proof}
\label{theorem:c_converges}
\end{theorem}
\medskip

Although our proofs rely on the continuity of $p$, we suspect a weaker version of this theorem is true.
We notice that in practice, this method estimates discrete variables fairly well, which can be thought of as delta functions in a continuous space.

\begin{figure}[!th]
    \centering
    \includegraphics[width=0.8\textwidth]{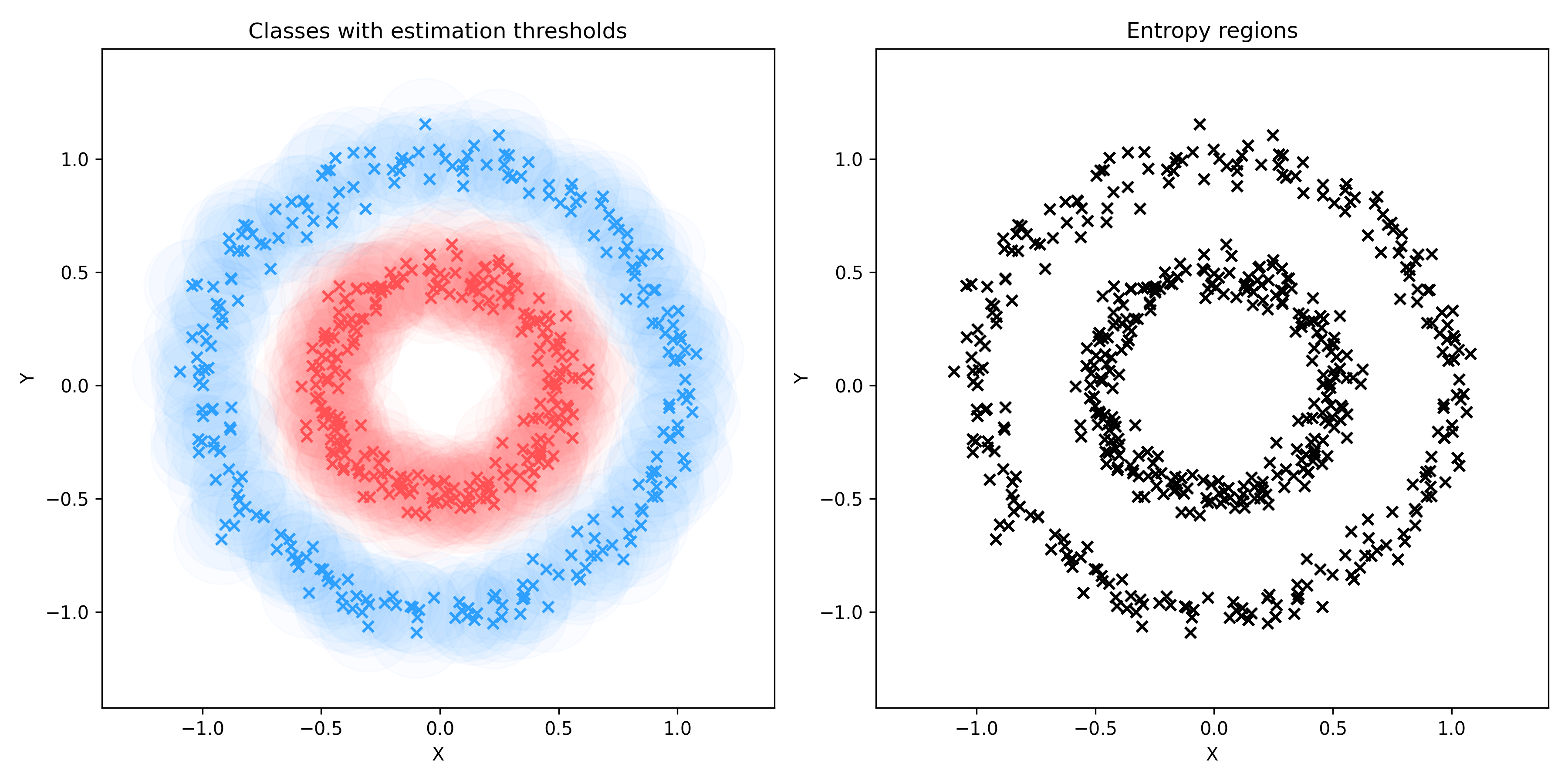}
	\includegraphics[width=0.8\textwidth]{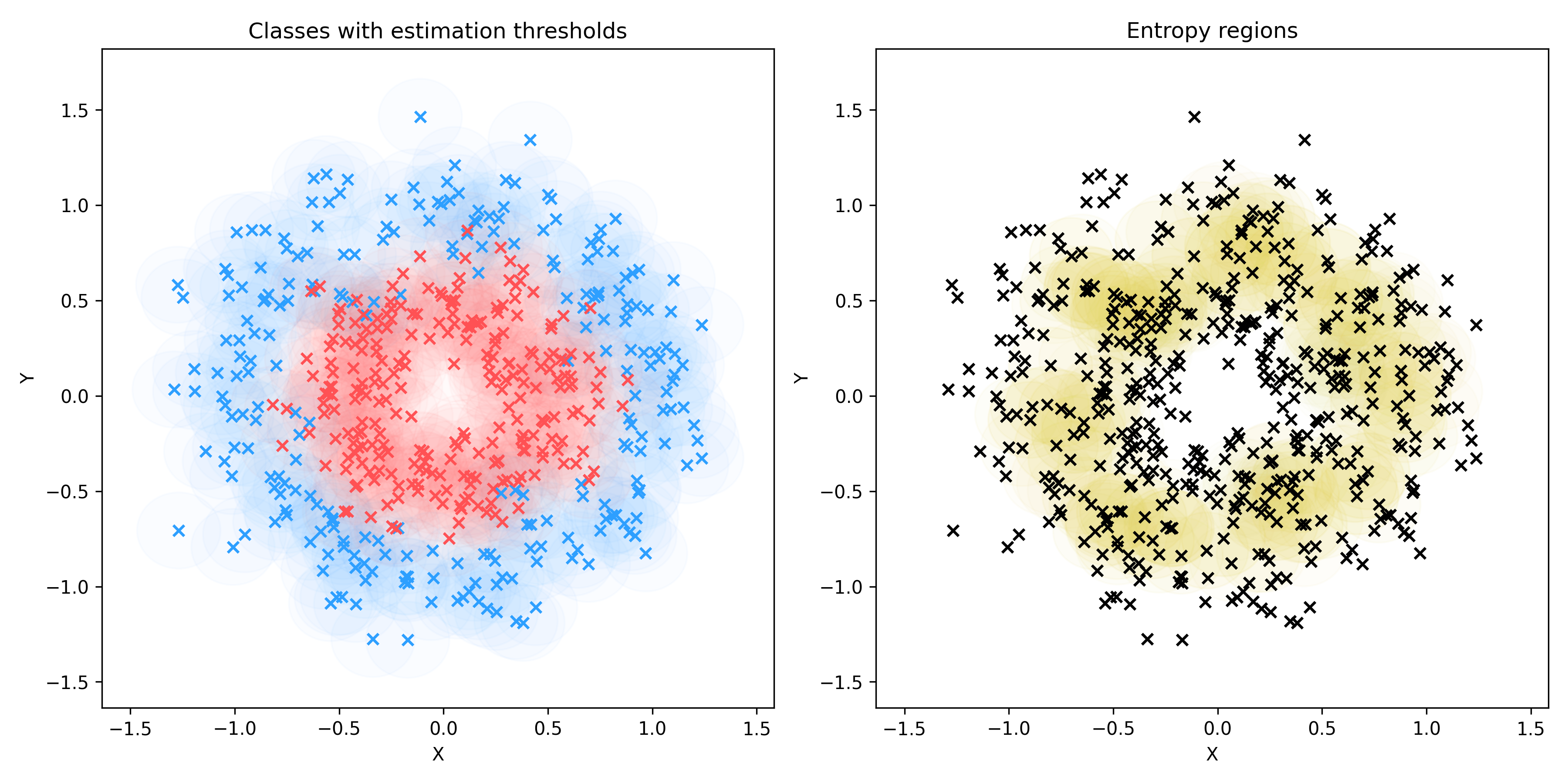}
    \caption{
        A visual example of the entropy estimation. 
        To the left we show two instances of two concentric circles with different levels of noise. 
        Each sample is indicated with an x marker, and a disk around the marker indicates the threshold for estimation.
        To the right, the corresponding entropy estimation is computed using the procedure outlined above. 
        Entropy is indicated in yellow, and the intensity of the color is proportional to the entropy value.
    }
    \label{fig:visual_example}
\end{figure}

Pictorially, it is easier to visualize the previous computation if we consider a simple example of two concentric circles, each representing a different class, as shown in Figure~\ref{fig:visual_example}.
Now, if we consider a disk around each sample, we can visualize which points affect the entropy estimation around that particular point. 
Further, we can use the same idea to illustrate the conflicting regions, i.e., regions with high entropy, giving a clear and intuitive picture of what our measure takes into consideration when computing the limit of the clasificability for the problem.

\textbf{}

\section{Not enough points}\label{sec5}

The previous methodology introduces one hyperparameter, the threshold distance we use to define the subsets $\mathcal{B}$.
As in the lower dimensionally fitting problems, indefinitely reducing the threshold $\theta$ value eventually leads to overestimating the classifibicability. 
It is a typical illusion.

In fact, if all classes have at least one continuous non-trivial random variable, $\widehat{\mathscr{C}_{D}}$ converges to one with high probability.

\medskip
\begin{lemma}
Let $\theta \in \mathbb{R}$ be some threshold and $D$ be some dataset composed of $\alpha$ classes $D_{\alpha} \subset D$, such that at least each class has at least one continuous non-trivial random variable. 
Then, the estimator $\widehat{\mathscr{C}_{D}}$ converges to one, with high probability, as $\theta \to 0$. In other words,
\begin{equation}
	 \lim_{\theta \to 0 } \widehat{\mathscr{C}_{D}} = 1 .
\end{equation}
\begin{proof}
Note that, with high probability, there are no two samples $\bar{x} \sim f_{\alpha}$ and $\bar{y} \sim f_{\beta}$ such that $\bar{x} = \bar{y}$ in $D$, since $D$ is finite and $\exists i, j$ such that $x_{i}$ and $y_{j}$ are continuously distributed.
Thus, with high probability, $\exists \epsilon > 0$ such that $\forall \prts{\bar{x}, \alpha} \in D, \mathcal{B} \prts{\bar{x}, \epsilon} = \brkt{\bar{x}}$.
Hence the $\forall \prts{\bar{x}, \alpha} \in D, \mathcal{H} \prts{ \bar{x}, \epsilon } = 0$.
\end{proof}
\label{lemma:class_converges_th_0}
\end{lemma}
\medskip

On the other hand, if the threshold $\theta$ is too large, we will understimate the limit of the classificability of the dataset.
When $\theta \to \infty$ we can show that $\widehat{\mathscr{C}_{D}}$ converges to the proportion of the most common class $\alpha$ in the dataset.

\medskip
\begin{lemma}
Let $\theta \in \mathbb{R}$ some threshold and $D$ be some dataset composed of $\alpha$ classes $D_{\alpha} \subset D$. Then, the estimator $\widehat{\mathscr{C}_{D}}$ converges to the proportion of the most common class as $\theta \to \infty$. In other words,
\begin{equation}
	 \lim_{\theta \to \infty } \widehat{\mathscr{C}_{D}} = \argmax{\alpha} \frac{ {\card{D_{\alpha}}} }{ \card{D} }.
\end{equation}
\begin{proof}
Note that when $\theta \to \infty$, $\mathcal{B}$ covers the entire dataset. Thus $\forall \alpha$, $\rho \prts{\bar{x},\alpha} = e^{1-\hat{p}\prts{\bar{x}, \gamma}}$ where $\gamma$ is the most represented class and $\hat{p} \prts{\bar{x}, \alpha}$ converges to the proportion of samples in class $\alpha$. Therefore, 
\begin{align}
\lim_{\theta \to \infty } \mathcal{H} \prts{ \bar{x}, \theta } &= 
	- \sum_{ \alpha } \frac{ {\card{D_{\alpha}}} }{ \card{D} } \log \prts{ \frac{ 1 }{ e^{1-\hat{p}\prts{\bar{x}, \gamma}} } } ,\\
\lim_{\theta \to \infty } \mathcal{H} \prts{ \bar{x}, \theta } &=  1 - \frac{ {\card{D_{\gamma}}} }{ \card{D} } .
\end{align}
Finally, with some algebraic manipulation, we arrive at the desired result.
\end{proof}
\label{lemma:class_converges_th_inf}
\end{lemma}
\medskip

This leads to a conflicting scenario: we need to use a larger $\theta$ to compensate for the fact that datasets are ubiquitously finite, but using a larger $\theta$ biases our estimate. 
Alternately, we can estimate the entropy using a fixed number of neighbors instead of a ball of a specific radius.
Under certain circumstances, this approach will also converge to the quantity of interest.
However, this approach has a similar limitation when we have finite datasets.

\medskip
\begin{lemma}
Let $k \in \mathbb{Z}$ some integer and $D$ be some dataset. Let $\mathcal{B} \prts{ \bar{x}, k }$ be the set of the $k$-closest neighbors of $\bar{x}$. Then, if we replace $\mathcal{B} \prts{ \bar{x}, \theta }$ with $\mathcal{B} \prts{ \bar{x}, k }$, the estimator $\hat{p}$ converges to $p$ as $\card{D} \to \infty$ and $0 \ll k \ll \card{D}$. In other words,
\begin{equation}
	 p \prts{\bar{x}, \alpha} = \lim_{ \begin{smallmatrix} 0 \ll k \ll \card{D} \\ \card{D} \to \infty \end{smallmatrix} } \hat{p} \prts{\bar{x}, k}.
\end{equation}
\begin{proof}
Since $\card{D} \to \infty$, $\exists \epsilon > 0$ such that $\card{ \mathcal{B} \prts{ \bar{x}, \epsilon } } = c$, with $0 \ll c \ll \card{D}$.
Let $k = c$, then  $\mathcal{B} \prts{ \bar{x}, k } = \mathcal{B} \prts{ \bar{x}, \epsilon }$. 
But by lemma \ref{lemma:p_converges}, $\hat{p}$ converges as $p$.
\end{proof}
\label{lemma:p_k_neighbors_converges}
\end{lemma}
\medskip
\begin{lemma}
Let $k \in \brkt{1,\dots,\card{D}}$ some integer and $D$ be some dataset. Let $\mathcal{B} \prts{ \bar{x}, k }$ be the set of the $k$-closest neighbors of $\bar{x}$. Then, if we replace $\mathcal{B} \prts{ \bar{x}, \theta }$ with $\mathcal{B} \prts{ \bar{x}, k }$, the estimator $\widehat{\mathscr{C}_{D}}$ converges to the proportion of the most common class as $k \to \card{D}$. In other words,

\begin{equation}
	 \lim_{k \to \card{D} } \widehat{\mathscr{C}_{D}} = \argmax{\alpha} \frac{ {\card{D_{\alpha}}} }{ \card{D} }.
\end{equation}

\begin{proof}
The proof is similar to lemma \ref{lemma:class_converges_th_inf}, since in the limit $\mathcal{B}$ also covers the entire dataset.
\end{proof}
\label{lemma:class_converges_k_neighbors_inf}
\end{lemma}
\medskip

\begin{figure}[!th]
    \centering
    \includegraphics[width=1\textwidth]{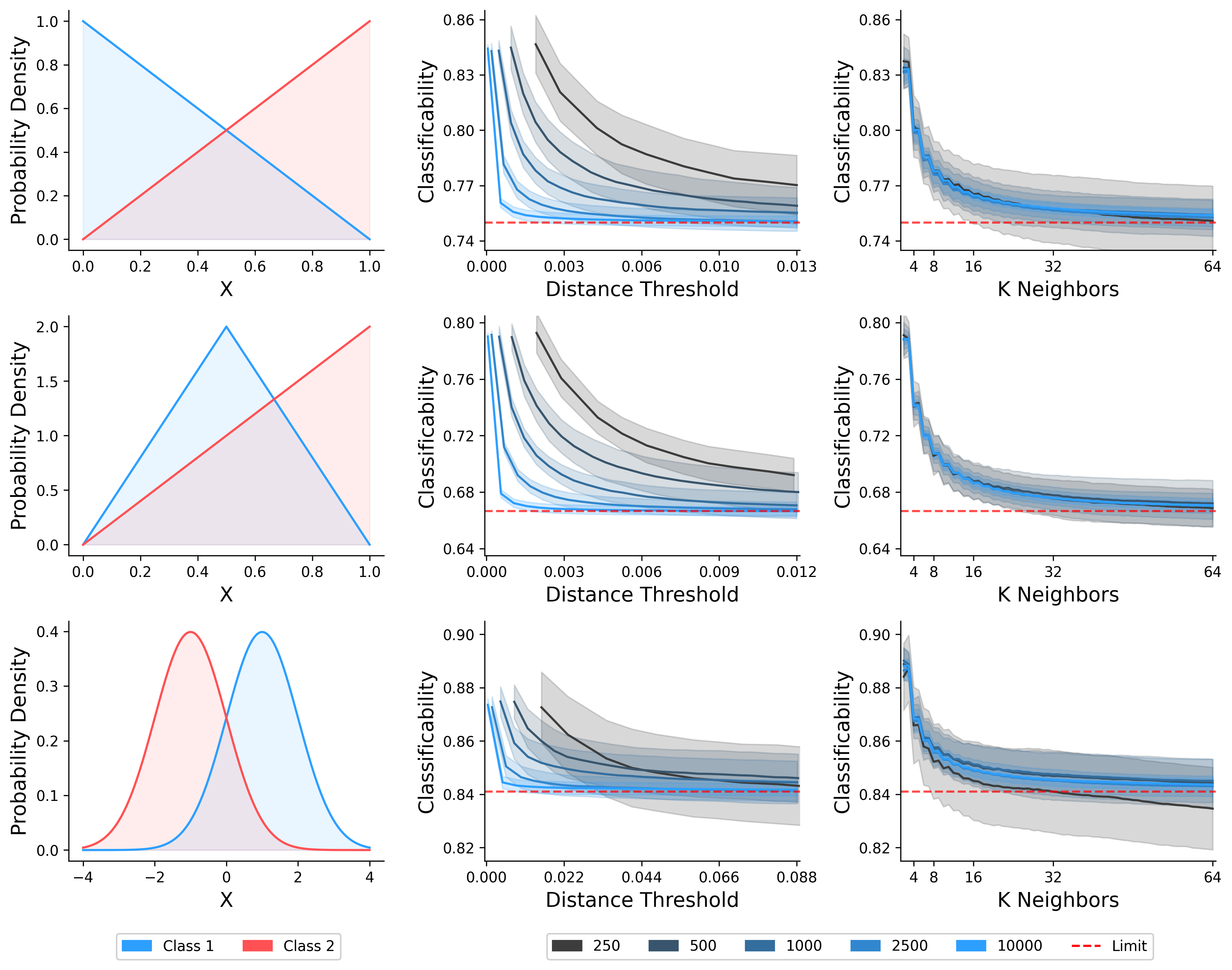}
    \caption{
    Three simple examples of the effect of having a finite dataset on the estimation of the classificability of the dataset. 
    Each row represents a dataset.
    The first column is the underlying distribution of each class on the dataset. 
    The second column showcases the effect of classificability as a function the threshold distance  $\theta$. 
    The third column is similar to the second one, but a fixed number of closest neighbors was selected instead of a fixed distance threshold to compute the estimates. 
    Each curve corresponds to datasets of different numbers of points per class. Shadows represent one s.d. computed from 25 samples (i.i.d. datasets). 
    The red line corresponds to the theoretical classificability limit for that dataset.
    Note that for these problems, the optimal model corresponds to a threshold model at the intersection of the PDF's.
    }
    \label{fig:finite_size}
\end{figure}

Figure~\ref{fig:finite_size} shows how our technique behaves on three simple datasets with two classes.
Distance thresholds were selected by computing the mean distance of points, regardless of the class, for several proportions of the dataset, ranging from 1\% to 5\% of the total size of the dataset.
Note that the size of the dataset heavily influences distance thresholds; for larger datasets, a small radius produces reasonable estimates, in contrast to smaller datasets, for which good estimates require a larger radius.
Still, the number of samples in the datasets does not significantly affect the k-neighbors approximation.
However, this approach tends to underestimate for a large number of neighbors.

\begin{figure}[!th]
    \centering
    \includegraphics[width=1\textwidth]{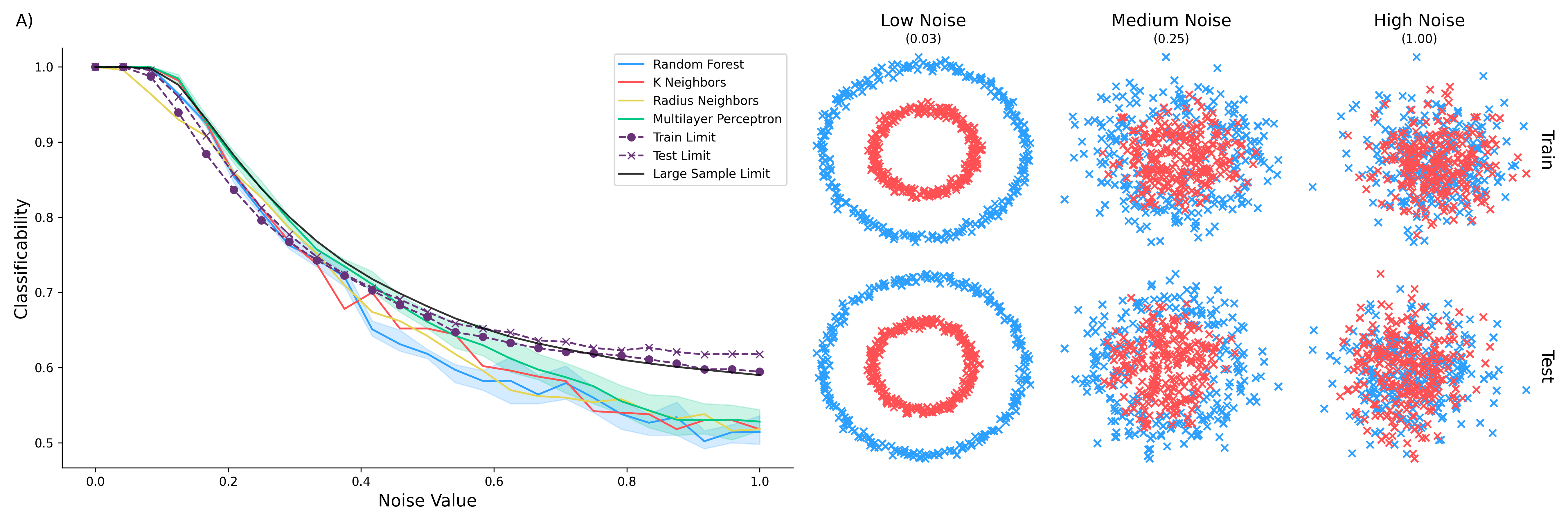}
	\includegraphics[width=1\textwidth]{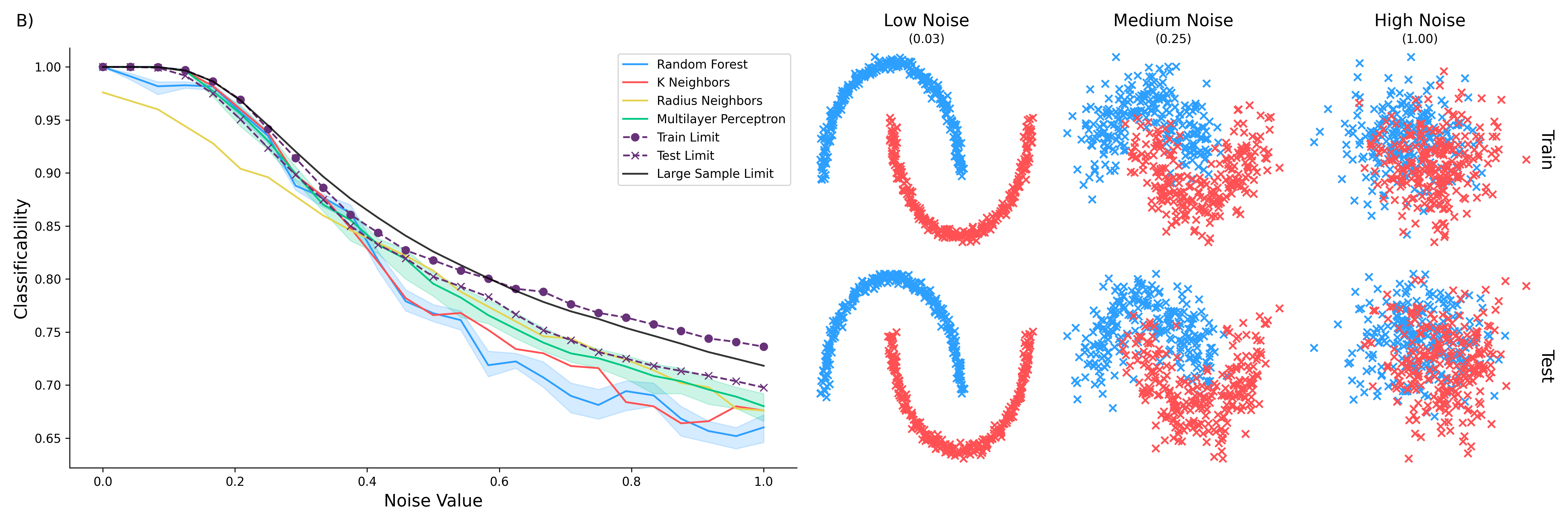}
	\includegraphics[width=1\textwidth]{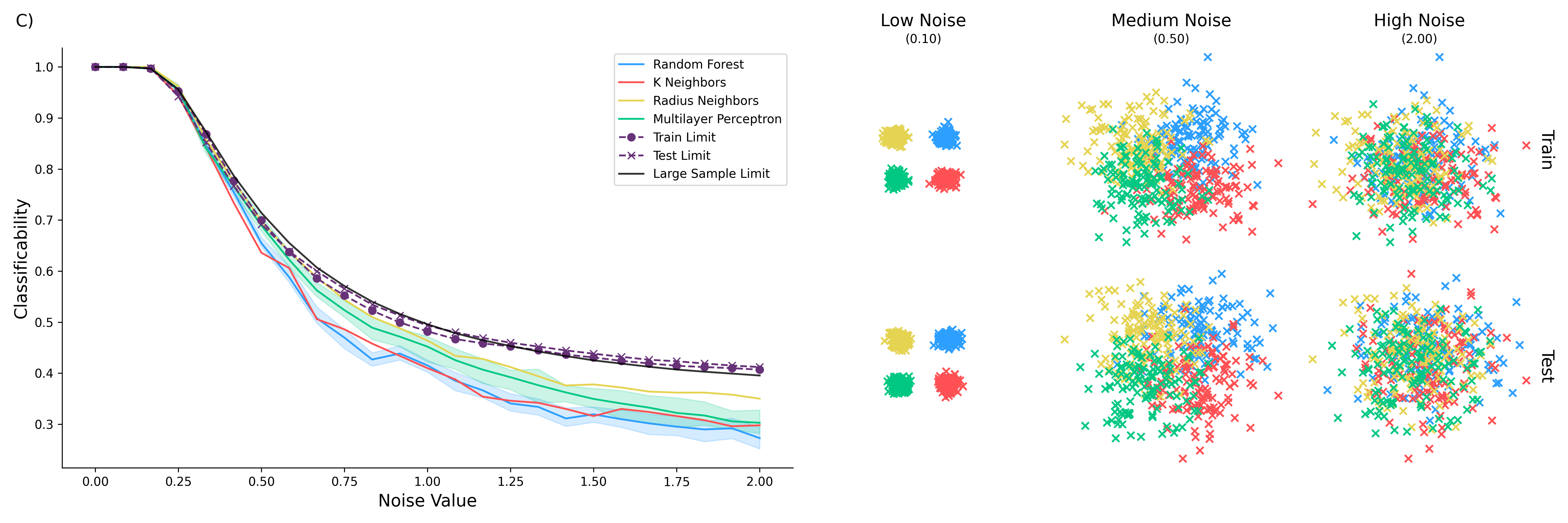}
    \caption{
    Classificability of three different datasets as a noise function: A) Concentric Circles, B) Two Moons, and C) Four Blobs.
    To the left, we show the performance of four different models: random forest, k-neighbors, radius neighbors, and a simple 3-layer neural network (128,64,32 units) with ReLU activation functions trained with Adam optimizer and L2 regularization (notice the changes in $y$ axis scales). 
    To the right, there is a depiction of how the datasets change as a function of noise.
    Note that the models were trained and tested in different datasets, although they used the same datasets for a particular noise level across all models. 
    Classificability limits were computed for the train and test datasets, 500 samples per noise level, using 16 neighbors.
    An additional classificability limit was computed using 25000 samples per noise level, using 32 neighbors.
    Note that our classificability estimation predictions are consistent with the overall behavior of the models.
    Small deviations are expected due to the finite size of the test datasets. 
    }
    \label{fig:simple_datasets}
\end{figure}

We evaluated the performance of our measure estimation in a set of controlled datasets and contrasted it with the empirical observation of different classifier models.
Our benchmark consisted of three 2D datasets, for which we can control the noise level of the dataset.
For each noise level, we sample two datasets of 500 points each: one for training and another for evaluation.
An estimation of the classificability was computed independently for the train and the test datasets. 
An additional dataset of 25,000 points was used for a high-resolution estimation of the classificability.
We trained four different types of models on each noise level using the same train and evaluation datasets: a random forest, a k-neighbor, a radius-neighbor, and a neural network.
For both the random forest and the neural network, we trained 25 different models with the same hyperparameters to account for the stochasticity of the training.
Figure~\ref{fig:simple_datasets} shows the classificabilities and actual accuracies of the trained models; we also show examples of the datasets at three noise levels to illustrate the effect of noise on the datasets. 

The comparison of the performance of various methods shown in Figure~\ref{fig:simple_datasets} leads us to some observations regarding the impact of the coordinate system on the classificability of a dataset. 
Our experiment uses an Euclidean coordinate system to describe the position of each data point. 
However, when splitting point categories of these particular data sets, the expectancy of a good classification performance is highly affected by the ``capacity'' of the coordinate reference system to evidence the different outcomes from different class points.  
A little experiment helps us to convey our argument. 
Focusing on the two moons' point distribution, we observe the two classes are easily separable by assessing a function correlating the radius and the angular portion of each point, suggesting the polar coordinate system is a better option to describe this specific type of point dispersion. 
The distribution of concentric points is an example of another interesting phenomenon: the points' angular position affects the separability of classes very little. 
Especially for small noise levels compared to the elongation of the elliptic distributions, this effect grows until the problem lowers its dimensionality and becomes a unidimensional problem solvable by observing only the radius. 

Even though it may sound obvious or a circular argument, we want to emphasize the importance of selecting an appropriate coordinate system to describe the dataset. 
At the same time, we recognize that identifying the underlying metric and/or topology of the space for a specific point distribution is an extremely difficult --- if not impossible --- problem to structure, especially when all ``real problems'' are derived from finite datasets. 

The neural network showed the general best classification ability out of the four models tested. 
We attribute the differences in this comparative behavior to the neural network's capacity to adapt to highly non-linear relationships among the parameters involved in the classification process. 
For low noise levels, the capacity of the coordinate system to synthesize non-linear relationships in the data space has little effect on the classificability model performance. 
Thus, we expect good performance of most classificability models for low-noise scenarios, as Figure~\ref{fig:simple_datasets} exhibits. 
This reasoning leads to hypothesizing the difference in performance between a classification model A and a neural network model B used as a reference, which indicates the potential improvement in the classificability achievable by adjusting non-linearity in the coordinates system used to depict the data space. 

Selecting the coordinate system to represent the data in a high-classificability data space is challenging. Hyper-dimensional spaces are intrinsically hard to intuit and interpret. Therefore, visualizing patterns is, at least, nearly impossible, and ``projecting'' which classification model performs better relies on a time-consuming testing process. A sharp change in the relative classificability experienced when the noise level increases from low to medium between one model and another is a strong signal of a better classification model for a given dataset.

\section{Too many classes}\label{sec6}

Next, we test our method in a collection of parametric datasets based on the Madelon dataset~\cite{madelon_dataset}.
The Madelon Dataset is a synthetic benchmark from the NIPS 2003 variable selection challenge. Initially implemented as a high-dimensional binary classification task featuring informative, redundant (linear combinations), and noise features, modern implementations allow for multiclass problem generation.
This dataset is designed to test feature selection and model robustness; its clustered data structure (hypercube vertices) and controlled label noise simulate real-world complexity, challenging algorithms to discern relevant features amid redundancy and stochasticity.

For the experiment, we build several datasets using 32, 64, 128, 256, and 512 raw features, from which 8, 16, 32, 64, and 128 are relevant features, respectively, and 4, 8, 16, 32, and 64 classes.
We generate five different datasets for each configuration, each with 25,000 samples. 
We preprocess each dataset with Linear Discriminant Analysis (LDA) to facilitate convergence to reasonable solutions with reduced computational power.
Classificability was computed after preprocessing to match the model performance more closely.
4-layer neural network models with L2 regularization were used to assess model performance.
Each model was trained for 100 epochs with 256, 128, and 64 hidden units, except for the case with 64 classes, for which we doubled the number of units per layer.

\begin{figure}[!th]
    \centering
    \includegraphics[width=0.99\textwidth]{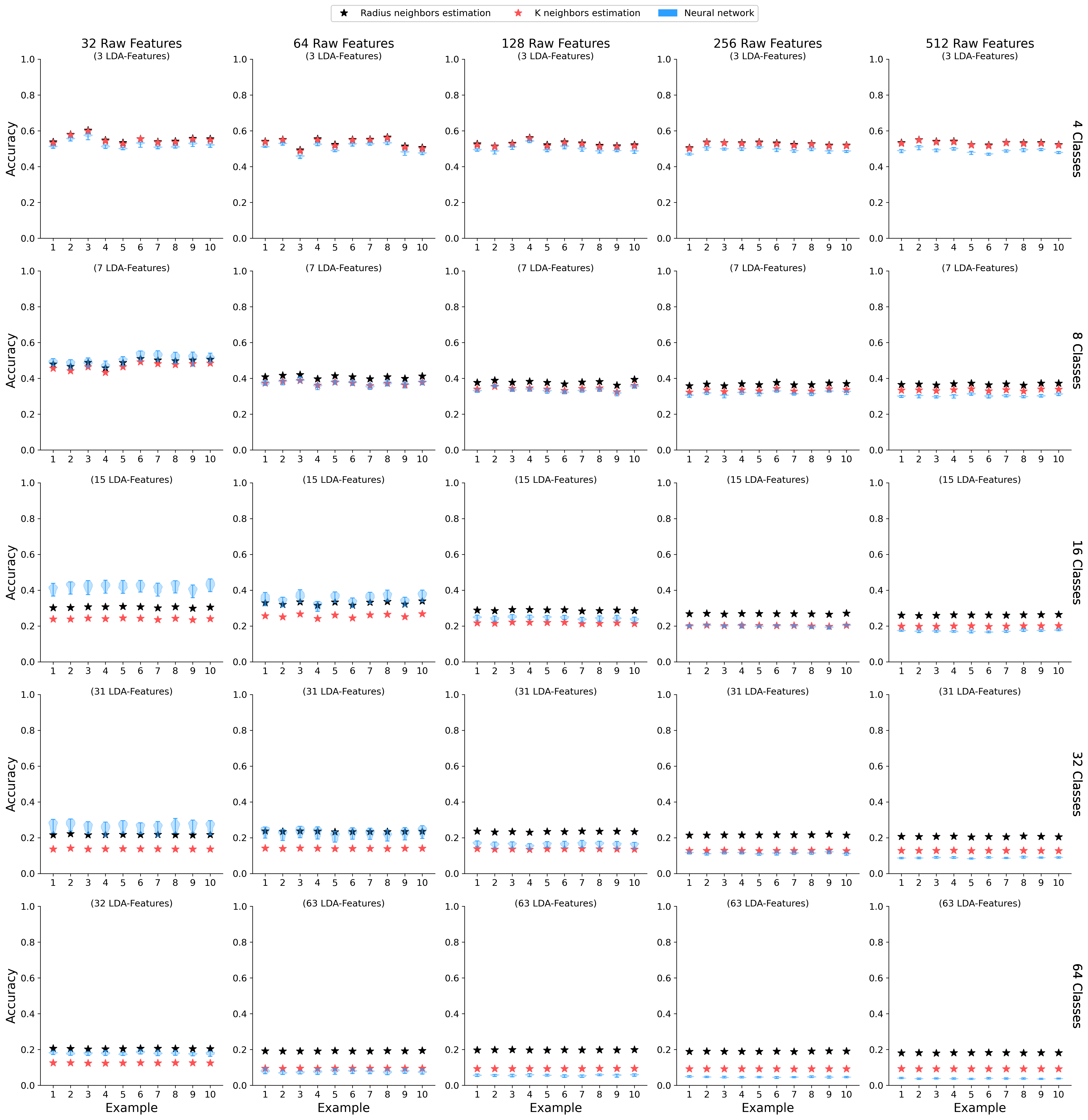}
    \caption{
    Classificability over several datasets with 32, 64, 128, 256, and 512 raw features and 4, 8, 16, 32, and 64 classes generated using the algorithm for the Madelon dataset~\cite{madelon_dataset}. 
    We generated five different datasets for each parameter configuration, each consisting of 25,000 samples. 
    Each dataset was preprocessed using LDA to ease model training. 
    Model performance is estimated from 5 different neural network models with L2 regularization.
    Classificability limit estimations are depicted with black stars (radius neighbors estimation) and red stars (k neighbors estimation). 
    Distance thresholds were computed as the mean distance between the closest 125 points, while 32 neighbors were used for the k neighbors estimation.
    The limit estimations for 4 and 8 classes tend to closely follow the trend of the performance of the neural network models, with a minor shift.
    This shift is highly likely to be introduced by the naive way in which the probabilities are estimated.
    Moreover, this shift is dependent on the hyperparameter of the model (distance threshold \& the number of neighbors), but for the sake of comparison, it was kept at a fixed value.
    We observe a degradation in the estimation quality as we increase the number of classes.
    Posterior experiments (data not shown) suggest that the real limit may be closer to the performance of the neural networks; however, it is still possible that the real limit may be higher.
    }
    \label{fig:madelon_datasets}
\end{figure}

Figure~\ref{fig:madelon_datasets} summarizes the experiment, depicting the classificability limit estimations with black and red stars, representing the threshold and the k-neighbors estimation, respectively.
Overall, our theoretical estimates predict the behaviour of the neural network models well for the experiments with 4 and 8 classes. 
We observe that our estimate matches, on a case basis, the performance pattern of the trained models with a minor shift. 
This shift is believed to be introduced by the rather naive approach used to estimate the probabilities around each point, which are sensible to the hyperparameter of the estimator. 
Our estimation seems sensitive to the subtleties of the datasets, at least for a small number of classes.
Overall, we observe that the k-neighbors-based estimations are consistent with the empirically observed model performance for the majority of the experiments.

For a larger number of classes, it is not clear whether our technique is excessively smoothing the estimate or indicates that there is a better strategy than the ones used by the neural networks, given their specifications.
Experiments conducted with a diverse set of models, weaker and stronger (data not shown), converge to a similar performance to the data in the plot, vouching for the first option, i.e., our estimation starts to dwindle for a large number of classes.
However, this is not strict evidence that this is, in fact, the limit of the data but rather of our ability to come up with better models. 
Nonetheless, depending on the downstream requirements, even our estimates for large number of classes can still provide actionable information for model / dataset building. 

\section{A small reality check}\label{sec7}

Until now, our results are entirely theoretical or based on synthetic datasets.
Thus, one may wonder if our estimations can address real-world problems.
Consequently, we train four different classifiers on several real-world datasets and contrast the model accuracy with our limit estimation.

Our previous results do not say anything about the metric of the space, but metrics do matter.
In the previous examples, we assumed an Euclidean metric. However, not every problem of interest to humans exists in an Euclidean space; we must remember that many real-world variables are categorical or binary.
This means that, in practice, we also need to consider the intrinsic metric of the space the data lives in to estimate the classificability correctly, e.g., fractal spaces or curved spaces.
Consider a U-shaped space in which each class lives on one of the tips of the U.
If we assume that the space is Euclidean, as we squish the U, the distance between classes will eventually be so small that we will underestimate the dataset's classificability. 

\begin{figure}[!th]
    \centering
    \includegraphics[width=1\textwidth]{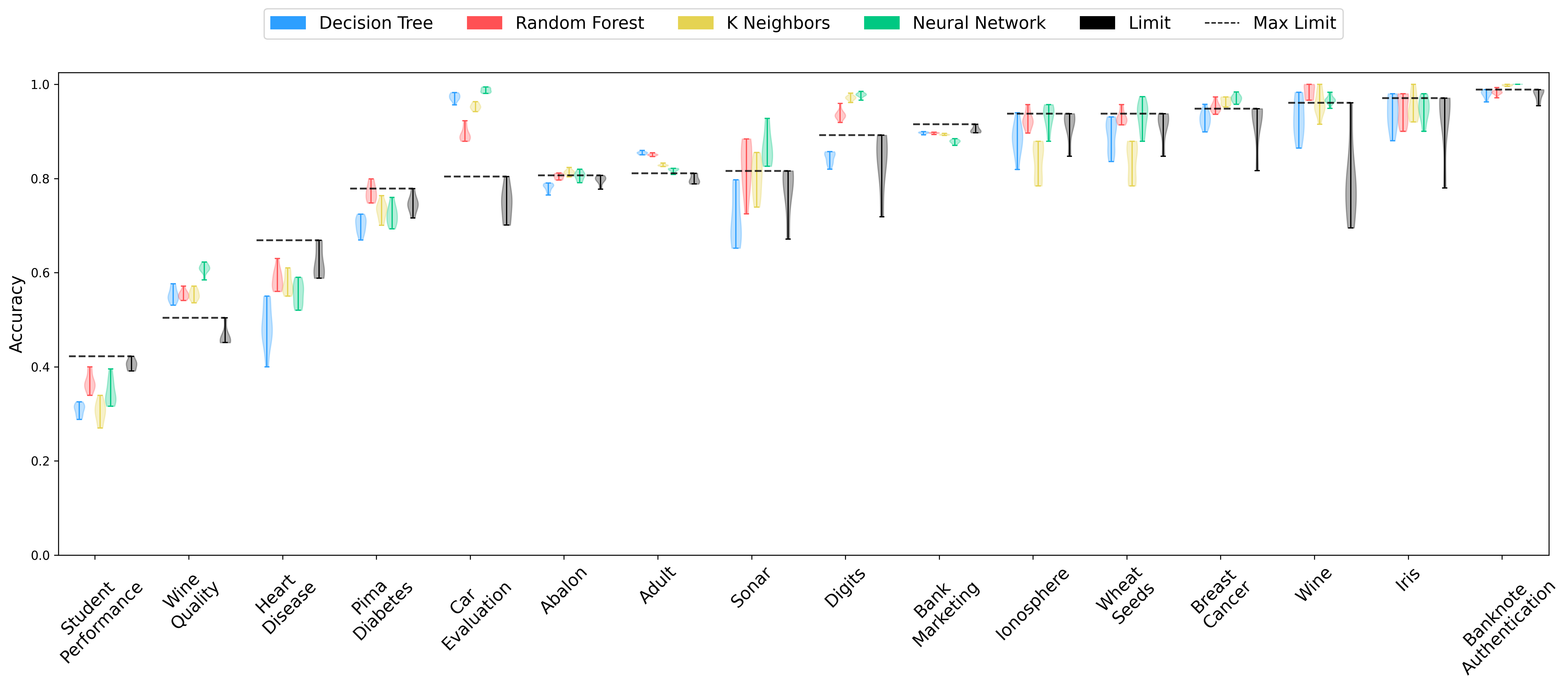}
    \caption{
    Classificability limit estimation for several real-world datasets.
    Classificability limits were computed using  1.5\% of the total dataset size as neighbors to each point. Due to the problems mentioned above, we clip the number of neighbors to a minimum of 6 and a maximum of 32.
    For comparison, we trained four different classifiers: decision tree, random forest, k-neighbors, and neural network.
    We used out-of-the-box models for all datasets with minor tweaks to the depth of the decision tree and random forest models. 
    All datasets were preprocessed with a standard scaler only.
    We performed a 4-fold cross-validation on each dataset to better assess the performance of each model.
    Results correspond to the mean accuracy with one standard deviation.
    Note that our estimations tend to be in agreement with the performance of the models.
    }
    \label{fig:reality_check}
\end{figure}

We selected a variety of datasets in several domains \footnote{https://archive.ics.uci.edu/datasets/}: healthcare, ecology, economy, physics, etc.
For the sake of simplicity, we stick to four out-of-the-box models: decision tree, random forest, k-neighbors, and neural network, with minor tweaks to their original configuration.
For each dataset, we trained each model on ten different stratified shuffles of the dataset to better assess the performance of each model, with a proportion of two to one for the train-test splitting.
We computed our limit estimation using different metrics:  L1, L2, Chebyshev, Hamming, Canberra, and Bryancurtis.
Note that the metric that minimizes the entropy (and maximizes the limit) is closer to the natural metric of the underlying distribution of the dataset. 
Results are shown in Figure~\ref{fig:reality_check}.

In general, our limit estimation is consistent with the accuracies observed for most datasets.
For the majority of the datasets, our estimation stays around the average accuracy of the top-performing model on each dataset.
Note that strict upper bounds were not observed for several problems, which in most cases can be explained by finite-size effects, both from model training (small datasets tend to produce high-accuracy models) and our estimation method.

As we highlighted above, metrics matter.
In some problems, the metric we choose can have a major impact on the estimation.
The Canberra metric~\cite{canberra}, a weighted version of the L1 metric, is consistently one of the metrics that minimizes the entropy.
This observation becomes more interesting when we note that these datasets contain many categorical and binary variables, for which an Euclidean metric is not ideal, but something like an L1 metric may work better.
The other metrics that also worked well, the Hamming or the Chebyshev metrics, are also more natural for categorical and binary variables than the Euclidean metric.

Take, for example, the Car Evaluation dataset~\cite{car_evaluation}: it is relatively easy to understand why our estimation is significantly lower than the models' performance.
First, we must note that all six dataset attributes are categorical variables.
This means that all data points lie in a 6-dimensional finite grid.
Second, the output labels for each point in the grid is deterministic and learnable by a tiny decision tree. 
Hence, the data is perfectly classifiable and easily learnable by all the models.
Finally, the dataset is relatively small, meaning there are not enough samples for each point in the grid.
In other words, we need to sample several neighboring points in the grid for each estimation we compute, and by the very nature of the dataset, it is highly likely that we are going to estimate erroneously the probability of most samples in the grid, especially around the frontier between classes. 
We suspect that most of our underestimations arise from a combined effect of finite size plus the grid-like topology of the space.

Another example is the Digits dataset, a small version of the more popular MNIST.
This dataset should contain enough points for a reasonable estimation; nonetheless, this is not the case, as seen in Figure~\ref{fig:reality_check}.
So, let us think about the topology of that space.  
The space of all possible images in grayscale is a nice space suitable for several of the metrics listed above. 
However, the topology of the subspace of all possible (readable) handwritten digits is likely to be not as nice as the space of all possible images, as it may even be discontinuous.
Therefore, the metrics we are using are not suitable for distance estimation in a space like this. 
Samples like 1 and 7 or 3 and 8 may end up being closer than they actually are in their natural space.
Correctly assessing the topology/metric of the space is probably the biggest challenge for our limit estimation.

To provide a more solid ground to the previous discussion, we performed a subsampling experiment similar to Figure~\ref{fig:why_example} using the four largest datasets.
These datasets ranged from a few thousands to a few dozens of thousands of samples. 
Each dataset was 10 times for each subsampling proportion, and 10 random forest were trained on each subsampling, totaling 100 models per proportion value.
Although random forest tends not to be as powerful as a well-implemented neural network, it performed well in these datasets as seen in Figure~\ref{fig:reality_check}, making it a good proxy for model performance. 

Furthermore, as we have mentioned earlier, our estimation tends to fall on the naive side. 
A simple improvement to our methodology is to refine the estimation of the local probabilities via a bootstrap without replacement, also known as jackknife \cite{jackknife}. 
We computed a bootstrapping maximum limit for the classificability using 80\% of the total samples and a total of 10 subsamples.

Similar to Figure~\ref{fig:why_example}, we can observe that models tend to be overly optimist at a low number of samples and tend to generalize poorly.
Moreover, even though a jackknife is still a pretty simple strategy for modern standards, this simple strategy already shows noticeable improvements in three out of the four experiments, indicated by a closer convergence to the intersection of training and testing accuracies.
This intersection between train/test was observed to empirically correspond to the classificability limit in many experiments using synthetic data (data not shown).

\begin{figure}[!th]
    \centering
    \includegraphics[width=1\textwidth]{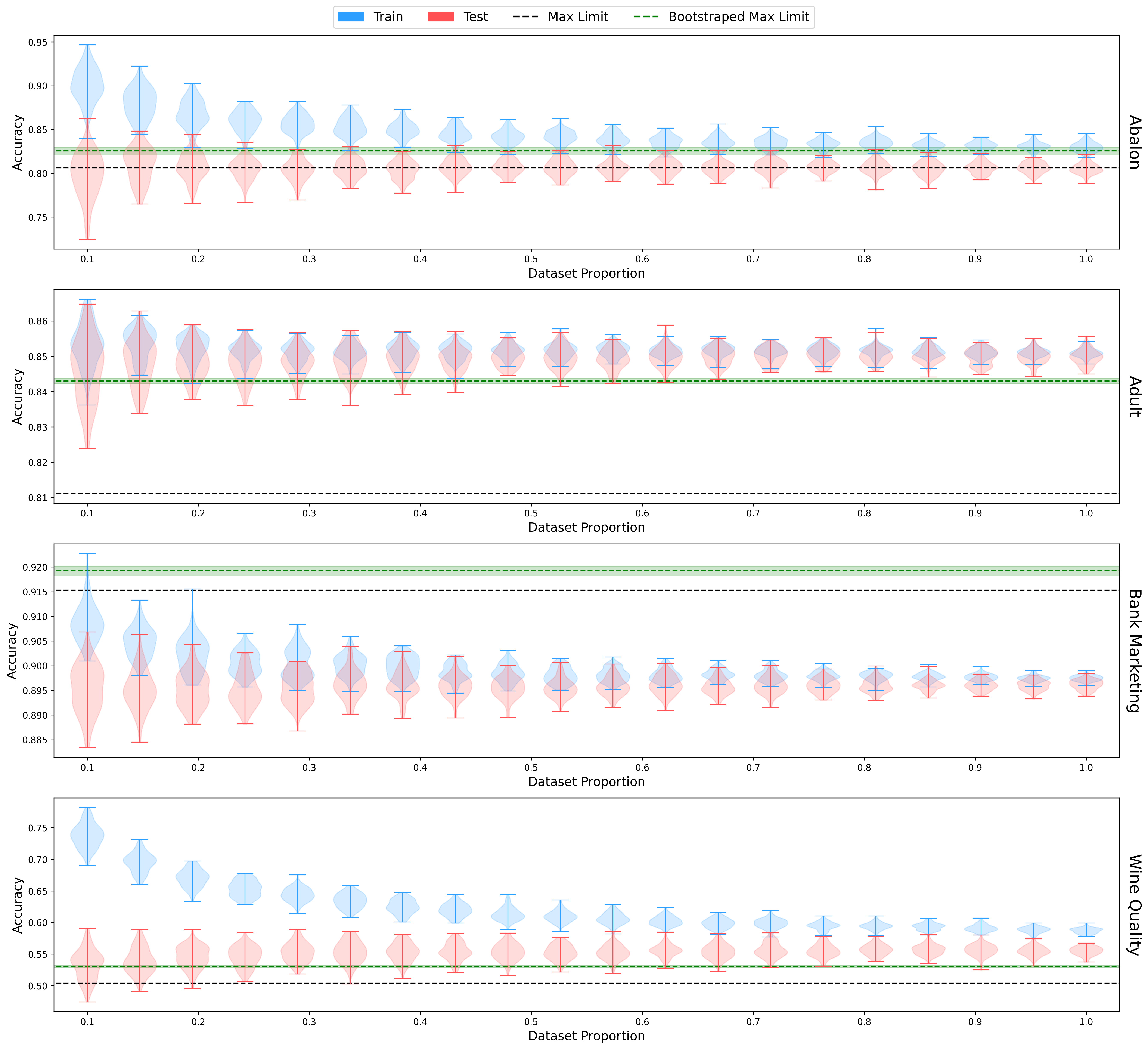}
    \caption{
    Subsampling limit behaviour of the Abalon, Adult, Bank Marketing, and Wine Quality datasets.
    Each dataset was drawn 10 times for each subsampling proportion, and 10 random forest were trained for each subsampling, for a total of 100 models per point.
    Max limit corresponds to the value shown in Figure~\ref{fig:reality_check}.
    Additionally, we computed a bootstraped value for max limit.
    Bootstraping is computed from a total of 10 subsamples of the dataset, using 80\% of the total samples.
    Errors indicate one std.
    A similar trend to Figure~\ref{fig:why_example} can be observed on real data. 
    }
    \label{fig:reality_subsampling}
\end{figure}


The number of classes in the model is an important parameter to account for. We can define over-classifying as the result of pretending to differentiate data points beyond the actual differences underlying the phenomena the data represent. Overfitting occurs when an over-dimensioned regression model is applied to a limited number of data points. When the number of dimensions, or namely random variables, approaches the number of data points, the regression model has enough degrees of freedom to produce a regression that is artificially close to the data points.

As an equivalent to overfitting in regression models, we focus on the overclassifying effect as it may appear in classification models. Over-classification may appear if the number of classes is not sufficiently larger than the number of data points. With this condition, the model may classify sets of points according to their position in the mathematically logical space, though it may not represent the actual and intrinsic properties describing the nature of the element each data point means.  The expression \ref{eq:spacesize} computes the number $N$ of potential classes in the classification model with $D$ dimensions and $r_d$, which is the resolution of dimension $d$, i.e., the number of different values upon the exclusive dimension $d$

\begin{equation}
    N = \prod_{d=1}^{D} r_{d}  
    \label{eq:spacesize},
\end{equation}

The complete set of combinations of dimensional parameter values forms the classification model space. Not all space classes have to be represented by data points in the dataset, so we qualify these as potential classes. However, $N$ is a good numeric reference for scaling the data volume required to produce reliable classification results. Considering the generally accepted rule of thumb that not less than 20 points are needed to distinguish a normal distribution from another, here we suggest a similar rule about the minimum limits of the number of data points $P \geq 20 N$. We are conscious the data may not uniformly populate the model space. Thus, some sub-spaces may only contain a few or even no points. On the other hand, other sub-spaces may contain a higher concentration of points forming clusters dispersed across the model space, thus compensating for the low-density effect and driving us to accept the criterion of a minimal data-point density for the complete model space. Finally, we emphasize the convenience of selecting the model dimensions to reduce their inter-dependence and, when possible, to diminish $N$, therefore putting the dataset farther away from the over-classification syndrome. 
Some of these rules are difficult to apply before explicitly executing the classification. Yet, keeping this guide in mind may be helpful in the iterative process of designing the classification model space.


\section{Conclusions}

We proposed a measure of the accuracy limit that a model may achieve in a classification problem.
However, this measure relies on knowing the density and normalization functions of the relative probabilities of each class.
Nonetheless, we showed that it is possible to estimate the limit of a classification problem from a dataset reasonably well.

Still, we suspect that there are ample opportunities for improvement, since our method for estimating the probability density functions is rather archaic. 
For instance, using importance sampling~\cite{importance_sampling} could potentially lead to significant improvements in estimating the entropy, leading to a finer estimate for the classification problem.
The present strategy for the entropy estimation could benefit from a bootstrapping strategy that considers the relevance of the other samples for the local entropy value.

It should be also mentioned that probably there are ``no free lunch'' theorems for classification~\cite{wolpert95no,wolpert1997no}. This would imply that there will be no single method that would classify data ``optimally'', nor even a single ``optimal'' classificability measure.

Estimating the intrinsic limits of a classification problem has the potential to help us build not only better classifiers but also better datasets.
Computing the classificability limit of a dataset, even if it is a proxy, can help us guide the search for meaningful attributes that better convey our intuitions about the classes we are aiming to describe. 
Perhaps one day, we will not need to wait until we have a really large dataset, in which thousands of hours from a few dozen souls were poured, just to discover that our models are not better than the flipping of a coin because our measurements of the problem were flawed.

Today, we are outsourcing more subtle and critical decisions to models such as random forest and neural networks, often built with the mindset that bigger is better. 
Then, we evaluate the quality of these classifiers by their ability to achieve a high value on a metric for a particular dataset. 
However, as we hope to have conveyed, optimizing for a specific dataset may differ from what we want in real-world applications. 
So we rely on splitting a dataset into increasingly complex pieces: train, test, validation, public test, private test, ultra-secret test, etc., hoping that, in the end, we can create a general model for the problem.
This approach works fine, but from time to time, the intrinsic biases that gave rise to the dataset are \textit{felt} by some sensitive programmers who may be able to pass it onto a model.
However, such approaches do not always generalize well and only apply to a particular subset of datasets, creating just an illusion of progress.  
The incapacity to push beyond an accuracy boundary is just one of the many limitations of artificial (or natural) intelligence: spurious correlations, contextual and semantic understanding, etc. 
We pursue models capable of generalizing, at least to some degree, the underlying problem, not the statistics of a dataset. 


\backmatter

\section*{Declarations}

\bmhead{Acknowledgements}
We thank Andrea Quintanilla for the fruitful discussions and partial revision of the mathematical analysis.

\bmhead{Funding}

\bmhead{Competing interests}
The authors declare that they have no competing interests.

\bmhead{Code availability}
Detailed code for the experiments is available on GitHub  \url{https://github.com/Nogarx/the-art-of-misclassification}

\bmhead{Data availability}
All synthetic datasets are available via the code on GitHub. All the real-world datasets used are available on \url{https://archive.ics.uci.edu/datasets/}.


\bibliography{references}

\end{document}